\documentclass[11pt,letterpaper]{article}
\usepackage{emnlp2016}
\usepackage{times}
\usepackage{latexsym}
\usepackage{amsmath}
\usepackage{color,soul}
\usepackage{booktabs}
\usepackage{threeparttable}
\usepackage{enumitem}
\usepackage{arydshln}

\newcommand{\my}[1]{}

\emnlpfinalcopy

\newcommand\cc[1]{\multicolumn{2}{c}{#1}}

\usepackage{array}
\newcolumntype{C}[1]{>{\centering\let\newline\\\arraybackslash\hspace{0pt}}m{#1}}

\title{A Factorized Model for Transitive Verbs in Compositional Distributional Semantics}

\author{Lilach Edelstein \\
	ELSC\\
	Hebrew University of Jerusalem \\
	{lilach.edelstein@mail.huji.ac.il} \And
	Roi Reichart\\
	Faculty of IE\&M\\
	Technion, IIT \\
	{roiri@ie.technion.ac.il} \\
}

\date{}
\begin{document}

\maketitle

\begin{abstract}
We present a factorized compositional distributional semantics model for the representation 
of transitive verb constructions. Our model first produces (subject, verb) and (verb, object) 
vector representations based on the similarity of the nouns in the construction to each of the 
nouns in the vocabulary and the tendency of these nouns 
to take the subject and object roles of the verb. These vectors are then combined into 
a  final (subject,verb,object) representation through simple vector operations.
On two established tasks for the transitive verb construction our model outperforms recent previous work. 
\end {abstract}

\section{Introduction}

In recent years, vector space models, deriving word meaning representations 
from word co-occurrence patterns in text, have become prominent in lexical semantics 
research \cite{turney2010frequency,Clark:12}.
Following this success recent attempts have been devoted to compositional
distributional semantics (CDS): combining the distributional word representations,
often in a syntax-driven fashion, to produce representations of phrases and sentences. 

Several tasks and techniques have been proposed for CDS. Some work aims at representing sentences 
that vary in length and structure mostly with neural network models
\cite{socher2012semantic,marelli2014sick,le2014distributed,marcobaronijointly}.
Another approach, which we take in this paper, is to focus on specific syntactic constructions.
At the expense of generality, this approach enables an in-depth investigation of a specific linguistic phenomenon.
\newcite{mitchell2008vector} proposed various additive and multiplicative operators for the 
combinations of word vectors, and applied them to intransitive verbs and their subjects. 
Recently, the categorical framework \cite{coecke2010mathematical,baroni2014frege} has been proposed, 
where each word is represented by a tensor whose order is determined by the categorical grammar
type of the word. For example, \newcite{baroni2010nouns} represent nouns by a vector, and adjectives
by matrices transforming one noun vector into another. 

In this paper we focus on the transitive verb construction, which recently attracts much attention.
In the categorical framework it is represented with 
a third order tensor that takes the noun vectors representing
the subject and object and returns a vector in the sentence space \cite{Grefenstette:13,polajnar2014reducing}. 
The main limitation of this approach is the excessive number of involved parameters. 
For example, a third-order tensor for a given transitive verb, mapping two 100-dimensional noun spaces
to a 100-dimensional sentence space, would have $100^3$ parameters in its full form.
Indeed, several recent works have tried to reduce the size of these models 
\cite{polajnar2014reducing,fried2015low} while others proposed matrix based representations 
\cite{polajnar2014reducing,milajevs2014evaluating,Paperno:14}.

We propose a {\it factorized model} for the representation of transitive verb constructions.
Given a subj-verb-obj $(s,v,o)$ construction, our model builds vector representations 
for the $(s,v)$ and the $(v,o)$ pairs, based on the similarity of $s$ and $o$ to each of the 
nouns in the vocabulary and the tendency of these nouns to take the subject and object roles of $v$.
The dimensionality of these $(s,v)$ and $(v,o)$ vectors (15701 in our case) equals to the number of 
nouns in the vocabulary that take the subject and object positions of $v$ frequently enough.
The $(s,v)$ and $(v,o)$ vectors are then combined to a final $(s,v,o)$ vector through simple vector operations. 
Our model outperforms recent previous work on two established tasks for the transitive verb construction 
\cite{Grefenstette:11b,kartsaklis2014study}.

\section{Model}
\label{sec:model}

The goal of our model is to generate vector representations (embeddings) for subject-verb-object $(s,v,o)$ constructions, 
where $s$ is the subject noun, $v$ is a verb and $o$ is the object noun. 
The model consists of two steps: 
{\bf (1)} Embed the $(s,v)$ and $(v,o)$ pairs based on co-occurrence statistics of the members of each pair 
with all other nouns in the vocabulary; and 
{\bf (2)} Combine the $(s,v)$ and $(v,o)$ representations to create a final $(s,v,o)$ embedding. 

Our model is a {\it factorized} model, generating an $(s,v,o)$ representation 
from its pair components ($(s,v)$ and $(v,o)$).
As such it is compact: the size of the $(s,v)$ and $(v,o)$ vectors is the number of nouns in 
the vocabulary that appear frequently enough both as subjects and as objects of $v$,  
and the $(s,v,o)$ vector is a simple derivation of these vectors.
In what follows we describe each of the above steps. 

\subsection{Pair Representations}

We represent the $(s,v)$ and $(v,o)$ pairs through the relations between the verb ($v$) 
and the noun ($s$ or $o$) with each other noun in the vocabulary. 
Particularly, we consider the tendency of each noun to come as a subject (for $(s,v)$) or object (for $(v,o)$)
of $v$, and the similarity of that noun to $s$ or $o$, respectively. By considering all the nouns in the vocabulary 
we get a smooth estimate of the tendency of the noun ($s$ or $o$) to take the subject (for $s$) or 
object (for $o$) position of $v$.
In what follows, we describe the representation in details for $(s,v)$ pairs. A very similar process is 
employed for $(v,o)$ pairs. 

For an $(s,v)$ pair we construct a vector representation whose size is the number of nouns that 
appear frequently enough at both subject and object positions in the training corpus. 
The k'th coordinate in this representation is given by:
\begin{equation}
\footnotesize
u^{subj}_{s,v}[k] = NVSubj(n_{k},v) \cdot NNSim(n_{k},s) \nonumber
\end{equation}
where $n_{k}$ is the k'th noun in the vocabulary, $NVSubj(x,y)$ reflects 
the tendency of the noun $x$ to be the subject of the verb $y$, and $NNSim(x,y)$ 
reflects the similarity between the nouns {\it x} and {\it y}. 
We next describe how $NVSubj(x,y)$ and $NNSim(x,y)$ are computed.

\paragraph{NVSubj}

For a noun $x$ and a verb $y$, we compute the positive point-wise mutual information (PPMI) of $x$ 
appearing as the subject of $y$, and the score for the $(x,y)$ pair is then given by:
\begin{multline}
\footnotesize
NVSubj(x,y) = PPMI^{subj}(x,y) =\\
max(0, log((P^{subj,verb}(x,y))/(P^{subj}(x)P^{verb}(y)))) \nonumber
\end{multline}
where $P^{subj,verb}(x,y)$ is the probability 
that a (subject,verb) pair in the corpus is $(x,y)$, 
and $P^{subj}(x)$ and $P^{verb}(y)$ are the probabilities that $x$ appears at the subject position 
of any verb in the corpus and that $y$ appears as a verb in the corpus, respectively. 

\paragraph{NNSim}

This score reflects the similarity between two nouns: $NNSim(x,y) = sim(x,y)$, 
where $sim(x,y)$ is any function that returns the similarity between its two word arguments.

We apply the same considerations for $(v,o)$ pairs: for the $k$-th coordinate, NVObj represents the tendency of 
$n_k$ to be an object of $v$, while NNSim represents the similarity between $n_k$ and $o$.

\subsection{$(s,v,o)$ Construction Representation}

Given the $(s,v)$ and $(v,o)$ representations described above, our next step is to combine them so that 
to get an effective representation of the $(s,v,o)$ triplet. 

We consider three combination methods. 
In the first two a single vector is constructed for $(s,v,o)$ through: (1) concatenation of the two vectors; 
and (2) coordination-wise multiplication:
\begin{multline}
\footnotesize
u^{(s,v,o)}[k] = u^{subj}_{s,v}[k] \cdot u^{obj}_{v,o}[k] =   \nonumber
NVSubj(n_{k},v) \cdot \\ NNSim(n_{k},s) \cdot 
NVObj(n_{k},v) \cdot NNSim(n_{k},o)
\end{multline}
That is, the $k$-th coordinate represents the similarity of $n_k$ to both $s$ and $o$, 
and its co-occurrence statistics with $v$ as both a subject and an object.
Under these two combination methods
the similarity score of two $(s,v,o)$ constructions is defined to be the cosine similarity 
between their vectors.

As an alternative, the third combination method keeps the $(s,v)$ 
and $(v,o)$ vectors as a representation of $(s,v,o)$.
The similarity between two constructions $(s,v,o)^1$ and $(s,v,o)^2$ is computed through the similarities between 
their components: 
$sim((s,v,o)^1, (s,v,o)^2) = \\ cosine(u_{(s,v)^1}, u_{(s,v)^2}) \cdot 
cosine(u_{(v,o)^1}, u_{(v,o)^2})$ 

\section{Experiments}

\paragraph{Data Preprocessing and Training}

We trained our models on the cleaned and tokenized Polyglot Wikipedia 
corpus \cite{AlRfou:2013conll},\footnote{https://sites.google.com/site/rmyeid/projects/polyglot} 
consisting of approximately 75M sentences and 1.5G word tokens. 
The corpora were POS-tagged with universal POS (UPOS) tags \cite{Petrov:2012lrec} using 
the TurboTagger \cite{Martins:2013acl},\footnote{http://www.cs.cmu.edu/~ark/TurboParser/} 
trained with default settings (SVM MIRA with 20 iterations) without any further parameter fine-tuning, 
on the {\sc train+dev} portion of the UD treebank. Following, the corpus was parsed with Universal Dependencies\footnote{We use UD so that 
in future work we can apply our model to other languages without major language-specific adaptations.} 
using the Mate parser v3.61 \cite{Bohnet:2010coling},\footnote{https://code.google.com/archive/p/mate-tools/} 
trained on the same UD treebank portion as the tagger and with default settings.

After parsing the corpus and before further statistics were collected, 
the corpus was lemattized to facilitate robust estimation. 
Therefore we also considered the lemmas of the words 
in our evaluation sets when computing an $(s,v,o)$ representation.
We extracted all $(n,v)$ and $(v,o)$ pairs based on dependency labels: 
a noun or a pronoun modifying a verb were considered its subject if their dependency arc 
is labeled "subj" or "nsubjpass", and its object if their dependency arc is labeled 
"dobj", "iobj", "nmod" or "xcomp".  
In order to reduce sparsity, our vocabulary contains only verbs that appear  at least 50 times in the corpus 
and nouns that appear at least 50 times both at subject and at object positions,  
a total of 6934 transitive verbs 
and 15701 nouns.

To compute the similarity between two nouns with $NNSim$, we trained the 
word2vec skip-gram model with negative sampling \cite{mikolov2013efficient} on our (unparsed) training corpus; 
context-window size was set to 5 and vector dimensionality to 200. 

\paragraph{Evaluation}

We evaluate the performance of our models on two well established tasks for transitive verb constructions. 
Both tasks require ranking of transitive sentence pairs for semantic similarity. 
The gold standard ranking is derived from similarity scores, on a 1-7 scale, provided by human evaluators.
The model ranking is evaluated against the gold standard ranking
using Spearman’s $\rho$.

The first task (GS11, \cite{Grefenstette:11b}) involves 
verb disambiguation: each of the 200 pairs in the dataset consists of two sentences that differ 
in their transitive verb, but share the same subject and object. 
For example, the members of the pair "(man, draw, sword), (man, attract, sword)" are less similar than
those of "(report, draw, attention), (report, attract, attention)".

The second task uses the transitive sentence similarity dataset (KS14, \cite{kartsaklis2014study}). 
This dataset consists of 108 subject-verb-object pairs, derived from 72 subject-verb-object triplets arranged into pairs. 
Unlike GS11, here each pair is composed of two triplets that differ in all three words.  
For example, the pair "(programme, offer, support), (service, provide, help)" is expected to 
get a higher similarity score compared to "(school, encourage, child), (employee, leave, company)".

In both tasks, we consider two different aggregation methods over the annotator scores of a pair:
(a) the human scores of each annotator are paired with the model scores without averaging, 
and a $\rho$ score is computed between the two vectors;
and (b) the human scores are first averaged, a human ranking 
is derived from the averaged scores and compared to the model ranking.
While the second method may seem more robust, it was not used in most previous works (see discussion in 
\newcite{mitchell2008vector}).

\begin{table}[]
\footnotesize
	\centering
	\begin{tabular}{l l c c c c}
		\toprule
		& 	\cc{Averaged} & 	\cc{Non-Averaged}\\
		\midrule
		 & GS11  & KS14 & GS11 & KS14\\
		
		\midrule
		w2v-sum & 0.280 & \textbf{0.760} &  0.210 & \textbf{0.60}\\ 
		w2v-all & 0.473 & 0.716 & 0.415 & 0.566 \\ 
		2014-best-non-simple  & 0.456 & 0.655 & -- & --\\
		2014-best-simple & 0.348 & 0.732 & -- & --\\ 
		2015-best-tensor  & 0.470 & 0.680 & 0.360 & 0.520\\
		2015-best-simple  & -- & 0.710 & 0.140 & 0.560\\ 
		w2v-PPMI-concat & 0.601 & 0.743 & \textbf{0.457} & 0.562\\
		w2v-PPMI-coor-mult & 0.598& 0.716 & 0.454 & 0.567\\
		w2v-PPMI-mult-score & \textbf{0.605} & 0.653 & 0.454 & 0.564\\
		\bottomrule	
	\end{tabular}
	\caption {\footnotesize Sprearman $\rho$ scores between model and human rankings for the GS11 
and the KS14 tasks with averaged and non-averaged human scores.
Top two models: baselines. Next four models: previous work. Bottom three models: this paper.}
\label{tab:title}
\vspace{-0.5cm}
\end{table}
 
 \paragraph{Models and Baselines}

We compare the results of our models, distinguished by the three combination methods: 
concatenation (w2v-PPMI-concat), coordination-wise multiplication 
(w2v-PPMI-coor-mult) and multiplication of the $(s,v)$ and $(v,o)$ scores (w2v-PPMI-mult-score).
We consider several baselines. In a first, simple baseline (w2v-sum), each $(s,v,o)$ construction is 
represented as the sum of 
the word2vec vectors of its words. This baseline, which corresponds to the unsupervised additive 
method of \newcite{mitchell2008vector}, captures the strength of the word2vec word level representations, 
ignoring word order and syntactic structure considerations.\footnote{We do not report results with coordination-wise 
multiplication of w2v vectors, as they lag behind other reported models.}

A second baseline (w2v-all) is similar to our method but the syntactic information is replaced with word similarity based 
on word2vec scores:
\begin{small}
\begin{multline}
u^{subj}_{s,v}[k] = NVSim(n_{k},v) \cdot NNSim(n_{k},s) \\
u^{obj}_{v,o}[k] =  NVSim(n_{k},v) \cdot NNSim(n_{k},o) \nonumber
\end{multline}
\end{small}
Where $NVSim(x,y)$ 
is the cosine similarity between the word2vec vectors of $x$ and $y$.
This method quantifies the importance syntax to our model.

Finally, we compare to state-of-the-art previous work:
(a) the most recent study on our tasks (\cite{fried2015low}, their table 1, 
{\it 2015-best-tensor}: their best model; {\it 2015-best-simple}: best result with additive 
or multiplicative combination as in \newcite{mitchell2008vector});
and (b) \newcite{milajevs2014evaluating} which performed exhaustive 
comparison of models and vector representations for our tasks (see their table 2,
{\it 2014-best-simple}: best result with additive or multiplicative combination; {\it 2014-best-non-simple}:
best result with tensor and matrix combinations based on 
\cite{Grefenstette:11a,Grefenstette:11b,kartsaklis:12,kartsaklis2014study}).\footnote{Note that   
*-tensor, *-simple and *-non-simple do not necessarily 
refer to the same model at all conditions: we pick the best result for each task and human scoring combination 
among the models in each of these model groups.}
 
\section{Results}

Results are presented in Table 1. 
Our models are superior on the GS11 task: 
the gaps between our best model and the best baseline are 13.2 and 4.2 $\rho$ points for the 
averaged and the non-averaged conditions, respectively. When comparing to the best previous work 
the gaps are 13.5 and 9.7 $\rho$ points, respectively.

For the KS14 task it is the simple w2v-sum baseline that performs best ($\rho = 0.76$ for averaged scores, 
$\rho = 0.6$ for non-averaged scores), but in both conditions it is one of our models that is second best 
(w2v-PPMI-concat with $\rho = 0.743$ for averaged scores, w2v-PPMI-coor-mult with 
$\rho = 0.567$ for non-averaged scores). Yet, in this task our gap from the baselines 
are smaller compared to GS11.

The superiority of a simple additive model for KS14 is in line with previously reported results.
While in KS14 the compared $(s,v,o)$ constructions do 
not overlap in their lexical content, 
in GS11 paired constructions differ only in the verb, hence requiring finer grained distinctions.
The relative difficulty of GS11 is also reflected by the lower scores all participating 
models achieve on this task. 

Importantly, while the w2v-sum excels on the KS14 task, its performance substantially degrade on the GS11 task 
where it achieves $\rho$ values of only 0.28 and 0.21 for the averaged and non-averaged cases respectively. 
Our models hence provide a sweet spot of good performance on both tasks.

Finally, the three variants of our model perform very similarly in three out of four test conditions. 
It is only for KS14 with averaged human scores that w2v-PPMI-mult-score lags 5.3 and 9 $\rho$ points 
behind w2v-PPMI-coor-mult and w2v-PPMI-concat, respectively. 
Hence, our model is flexible with respect to combination method selection.

\section{Conclusions}

We presented a  factorized model for $(s,v,o)$ embeddings which provides a simple and compact 
alternative to existing methods, and showed that it excels on two recent CDS tasks. 
In future work we intend to extend our model so that it accounts for more complex
syntactic constructions, such as, e.g., ditransitive verb constructions and constructions that include adjectives and adverbs. 

\bibliography{A Factorized Model for Transitive Verbs in Compositional Distributional Semantics}
\bibliographystyle{A Factorized Model for Transitive Verbs in Compositional Distributional Semantics}

\end{document}